\documentclass[pdflatex]{nolta}
\usepackage{mathtools}
\usepackage{bm}
\newtheorem{proposition}{Proposition}

\Vol{17}%
\No{1}%
\Year{2026}%
\Month{1}%
\PaperID{2026ENP0886}

\title{Geometry of learning dynamics: Gradient descent versus
  natural gradient on the ridge of optimization}

\AUTHOR{
\author{Akira Tamamori}{1}\orcid{0009-0000-8893-0058}
}

\AFFILIATE{ \affiliate{Faculty of Information Science, Aichi
    Institute of Technology\\1247 Yachigusa, Yakusa-cho, Toyota-shi, Aichi 470-0392, Japan}{1} }

\received{10}{27}{20XX}
\revised{12}{29}{20XX}
\published{7}{1}{20XX}

\begin{document}

\begin{abstract}
  High-capacity associative memories based on Kernel Logistic
  Regression (KLR) exhibit a \textit{Ridge of Optimization}
  characterized by extreme stability and a highly skewed weight
  spectrum. However, the dynamical process by which learning converges
  to this critical regime has remained unclear. This paper provides a
  geometric analysis of the learning trajectories on the statistical
  manifold of a KLR-trained Hopfield network. By comparing the paths
  of Gradient Descent (GD) and Natural Gradient Descent (NGD), we
  elucidate the mechanisms governing the optimization process. Our
  analysis reveals that learning on the Ridge proceeds in two distinct
  phases. We show that the extreme curvature of the Ridge causes
  standard GD to follow a highly oscillatory, non-geodesic path. In
  stark contrast, NGD explicitly corrects for this geometry, following
  the ideal geodesic path and completely overcoming the instabilities
  faced by GD. We demonstrate experimentally that NGD not only
  converges significantly faster but also achieves a solution with
  superior generalization performance. These results establish that
  the highly structured geometry of the Ridge is optimally suited for
  information-geometric optimization, providing a new perspective on
  the interplay between learning dynamics and emergent representation
  geometry.
\end{abstract}
\begin{keywords}
  kernel associative memory, learning dynamics, information geometry,
  natural gradient descent, edge of stability
\end{keywords}

\maketitle

\section{Introduction}
\label{sec:introduction}

High-capacity associative memories, realized through Kernel Logistic
Regression (KLR) trained Hopfield networks, have demonstrated
exceptional storage and retrieval performance, significantly exceeding
classical theoretical limits~\cite{tamamori2025, tamamori_nolta_a}. Our previous
analyses have characterized the static geometric properties of the
attractors in these networks, identifying a specific hyperparameter
regime termed the \textit{Ridge of Optimization.} On this Ridge, the
network achieves maximal stability by self-organizing its weight
spectrum into a highly concentrated, or ``L-shaped,''
distribution~\cite{tamamori_nolta_b}.

Despite a clear understanding of these final, converged states, a
fundamental question remains unanswered: \textit{how} do the learning
dynamics navigate the high-dimensional parameter space to reach this
specific, highly structured regime? The trajectory of the optimization
process, which dictates both the speed of convergence and the quality
of the final solution, has not yet been geometrically
analyzed. Understanding this dynamical pathway is crucial for
developing more efficient learning algorithms and for uncovering the
general principles that govern self-organization in such systems.

This paper provides a geometric analysis of the learning trajectories
in KLR-trained Hopfield networks. We investigate the optimization path
on a statistical manifold equipped with the Fisher Information Matrix
(FIM) as a Riemannian metric. By comparing the dynamics of standard
Gradient Descent (GD) with those of Natural Gradient Descent
(NGD)~\cite{Amari1998}, which explicitly accounts for the manifold's
curvature, we elucidate the geometric mechanisms that guide the
learning process.

Our main contributions are as follows:
\begin{enumerate}
\item We demonstrate that learning on the Ridge proceeds in two
  distinct phases: an initial, rapid phase where the dominant spectral
  mode of the FIM is established, followed by a prolonged fine-tuning
  phase for the remaining modes.
    
\item We visualize the learning trajectories in a projected eigenspace
  of the FIM. We show that the standard GD trajectory is severely
  constrained by the manifold's extreme curvature, resulting in highly
  oscillatory, non-geodesic paths that over-shoot the principal 
  direction. In stark contrast, the NGD trajectory follows the ideal 
  $e$-geodesic (straight line) directly to the optimum.
    
\item We compare the convergence speed and generalization performance
  of GD and NGD. We find that NGD, by actively correcting for the 
  manifold's curvature, entirely bypasses the instabilities faced by GD. 
  Consequently, NGD not only converges significantly faster but also 
  achieves a lower final validation loss, indicating superior generalization.
\end{enumerate}

These findings suggest that the specific geometry of the Ridge of
Optimization is uniquely suited to information-geometric optimization
methods like NGD. The remainder of this paper is organized as
follows. Section~\ref{sec:related_work} reviews related
work. Section~\ref{sec:methods} outlines the geometric framework and
optimization methods. Section~\ref{sec:results} presents the
experimental results on learning trajectories and
performance. Finally, Section~\ref{sec:discussion} discusses the
implications, and Section~\ref{sec:conclusion} concludes the paper.

\section{Related Work}
\label{sec:related_work}

Our research is situated at the intersection of information geometry,
optimization theory, and the dynamical analysis of neural networks.

\subsection{Information Geometry and Optimization}
The geometric structure of statistical models has been extensively
studied within the framework of Information Geometry, pioneered by
Amari~\cite{Amari2016}. A key insight from this field is that the
standard gradient descent follows the steepest descent direction in
the Euclidean parameter space, which may not be optimal on a curved
statistical manifold. The Natural Gradient Descent (NGD), proposed by
Amari~\cite{Amari1998}, corrects for the manifold's curvature using
the FIM, and is known to achieve optimal asymptotic performance. The
relationship between NGD and other optimization methods, such as
mirror descent, has been a rich area of
research~\cite{Raskutti2015}. While NGD is theoretically appealing,
its practical application is often hindered by the computational cost
of inverting the FIM. Our work provides a concrete experimental
validation of NGD's superiority in a non-trivial, high-dimensional
setting (the Ridge), motivating further research into efficient
approximations of the natural gradient.

\subsection{Learning Trajectories in Deep Neural Networks}
The analysis of learning trajectories has recently gained significant
attention in the deep learning community. Much of this work has
focused on the ``lazy training'' regime, where networks operate
similarly to fixed kernel machines, a behavior characterized by the
Neural Tangent Kernel~\cite{Jacot2018}. In this regime, the learning
dynamics are relatively simple.  However, for models operating beyond
this lazy regime, the interplay between learning dynamics and the loss
landscape is far more complex. A central topic in modern optimization
theory is the role of curvature and ``sharpness'' in determining
generalization. While conventional wisdom suggests that GD implicitly
favors ``flat minima'' to achieve better
generalization~\cite{Keskar2017}, the landscape of over-parameterized
networks often exhibits a highly skewed Hessian spectrum, dominated by
a few large outlier eigenvalues~\cite{Sagun2018}.

This extreme curvature leads to phenomena such as the \textit{Edge of
  Stability}~\cite{Cohen2021}, where GD trajectories are forced to
oscillate near sharp regions rather than settling into the absolute
minimum. Our analysis of KLR networks on the Ridge contributes
directly to this line of inquiry. By comparing GD's severe oscillatory
behavior with NGD's stable convergence, we provide a clear geometric
picture of how learning algorithms navigate these pathological,
high-curvature landscapes. Furthermore, our results offer a new
perspective on the ``flat minima'' debate, demonstrating that in
highly structured problems like associative memory, optimal
generalization can indeed be achieved in an extremely sharp basin,
provided the optimization method respects the intrinsic geometry of
the manifold.

\section{Model and Geometric Framework}
\label{sec:methods}

This section briefly reviews the Kernel Logistic Regression (KLR)
Hopfield network model and the geometric framework of Information
Geometry, which provides the foundation for our analysis of learning
dynamics.

\subsection{Kernel Logistic Regression Hopfield Network}
We consider a network of $N$ bipolar neurons,
$\bm{s} \in \{-1, 1\}^N$, trained to store $P$ random patterns
$\{\boldsymbol{\xi}^\mu\}_{\mu=1}^P$. The network's state evolves
based on the input potential $h_i(\bm{s})$, determined by the dual
variables $\alpha_{\mu i}$ and the RBF kernel $K(\cdot, \cdot)$:
\begin{equation}
h_i(\bm{s}) = \sum_{\mu=1}^P \alpha_{\mu i} K(\bm{s}, \boldsymbol{\xi}^\mu).
\end{equation}
The dual variables $\boldsymbol{\alpha}$ are learned by minimizing an
$L_2$-regularized negative log-likelihood objective function. For
simplicity, we focus on the trajectory of the weights
$\boldsymbol{\alpha} = [\alpha_{1}, \dots, \alpha_{P}]^\top$ for a
single representative neuron. The objective function
$L(\boldsymbol{\alpha})$ to be minimized is given by:
\begin{align}
  L(\boldsymbol{\alpha}) &= - \sum_{\mu=1}^{P} \left[ y_{\mu} \log(\sigma(h(\boldsymbol{\xi}^\mu)))\right. \nonumber\\
                         &\quad\quad\quad\quad+ \left.(1 - y_{\mu}) \log(1 - \sigma(h(\boldsymbol{\xi}^\mu))) \right] \nonumber \\
  & \quad+ \frac{\lambda}{2}\boldsymbol{\alpha}^{\top} \bm{K}\boldsymbol{\alpha}, 
  \label{eq:loss_function}
\end{align}
where $y_\mu \in \{0, 1\}$ is the target bit corresponding to pattern
$\boldsymbol{\xi}^\mu$, $\sigma(z) = 1/(1+e^{-z})$ is the logistic
sigmoid function, and $\lambda$ is the weight decay parameter. This
optimization is performed via full-batch Gradient Descent (GD).

\subsection{Statistical Manifold and Fisher Information}
The KLR model can be viewed as a parametric statistical model where
each set of weights $\boldsymbol{\alpha}$ corresponds to a probability
distribution $p(\cdot; \boldsymbol{\alpha})$ over the output
space. The set of all such distributions forms a statistical manifold,
whose intrinsic geometry is defined by the Fisher Information Matrix
(FIM), $G(\boldsymbol{\alpha})$~\cite{Amari2016}. For a single neuron,
the FIM is given by:
\begin{equation}
  G_{\mu\nu}(\boldsymbol{\alpha}) = \mathbb{E} \left[ \frac{\partial \log p}{\partial \alpha_\mu} \frac{\partial \log p}{\partial \alpha_\nu} \right].
\end{equation}
In the KLR context, this can be expressed in terms of the kernel Gram
matrix $\bm{K}$ and a diagonal matrix $\bm{D}$ containing the
prediction variances~\cite{tamamori_nolta_b}:
\begin{equation}
  G(\boldsymbol{\alpha}) = \bm{K} \bm{D}(\boldsymbol{\alpha}) \bm{K},
\end{equation}
where $\bm{D}(\boldsymbol{\alpha})$ is a diagonal matrix with
entries $D_{\mu\mu} = p_\mu(1 - p_\mu)$, representing the variance
of the prediction for pattern $\boldsymbol{\xi}^\mu$.  The FIM acts
as a Riemannian metric tensor, defining the notion of distance and
curvature on the statistical manifold.

Furthermore, information geometry introduces a dualistic affine
structure to the manifold, characterized by the exponential ($e$-)
connection and the mixture~($m$-) connection~\cite{Amari2016}. Since
the KLR model is based on a logistic function, which corresponds to a
Bernoulli distribution, it is a member of the exponential family.
Consequently, the natural parameters $\boldsymbol{\alpha}$ form an
$e$-flat coordinate system, and an \textbf{$e$-geodesic}, which
represents the ``straightest'' path connecting two probability
distributions with respect to the $e$-connection, corresponds simply
to a straight line in the Euclidean parameter space
$\boldsymbol{\alpha}$.  Conversely, the $m$-geodesic is a straight
line in the space of expectation parameters (the output
probabilities). In our trajectory analysis, we use the $e$-geodesic
(the straight line from the initial weights $\boldsymbol{\alpha}_0$ to
the optimal weights $\boldsymbol{\alpha}^*$) as the theoretical
geometric baseline for the most direct path in the parameter space.

\subsection{The Ridge of Optimization}
\label{sec:ridge_def}

In this study, we focus our analysis on a specific hyperparameter
regime, which we term the \textit{Ridge of Optimization.} To provide a
precise geometric context, we formally define this Ridge $\mathcal{R}$
based on the spectral properties of the FIM. Let
$\lambda_{\max}(G(\boldsymbol{\alpha}^*))$ be the largest eigenvalue
of the FIM evaluated at the converged weights
$\boldsymbol{\alpha}^*$. The Ridge is defined as the locus in the
hyperparameter space $(\gamma, P/N)$ where this local curvature is
maximized:
\begin{align}
  \mathcal{R} &= \{ (\gamma, P/N) \mid \lambda_{\max}(G(\boldsymbol{\alpha}^*)) \text{ is locally maximized} \}.
  \label{eq:ridge_definition}
\end{align}
We acknowledge that this is an \textit{a posteriori} definition, as it
depends on the converged solution $\boldsymbol{\alpha}^*$. However,
our previous extensive phase diagram analysis~\cite{tamamori_nolta_a}
demonstrated that the location of this Ridge is highly predictable and
structurally stable across different random initializations. This
region coincides precisely with a state of \textit{Spectral
  Concentration}, characterized by a highly skewed eigenvalue
hierarchy ($\lambda_1 \gg \lambda_2 \ge \dots > 0$).  Unless otherwise
stated, our experiments are conducted on representative points along
this pre-identified Ridge (e.g., $\gamma=0.02, P/N=2.0$) to
investigate the unique optimization dynamics that unfold in this
extreme-curvature environment.

\subsection{Natural Gradient Descent}
While GD follows the steepest descent direction in the flat Euclidean
space of parameters, the Natural Gradient Descent (NGD) follows the
steepest descent on the curved statistical
manifold~\cite{Amari1998}. The NGD update direction is obtained by
pre-conditioning the standard gradient with the inverse of the FIM:
\begin{equation}
    \tilde{\nabla} L(\boldsymbol{\alpha}) = G(\boldsymbol{\alpha})^{-1} \nabla L(\boldsymbol{\alpha}).
\end{equation}
This update step is invariant to re-parameterizations of the model and
is known to be asymptotically optimal. By comparing the trajectories
generated by GD and NGD, we can analyze how the manifold's curvature
influences the learning process. For numerical stability when
inverting the FIM, we use a damped inverse
$(G + \epsilon \bm{I})^{-1}$.

\subsection{Experimental Setup}
Unless otherwise specified, our experiments are conducted with a
network of $N=50$ neurons storing $P=100$ random patterns
($P/N=2.0$). We compare a representative ``Ridge'' regime
$(\gamma=0.02)$ with a ``Local'' regime $(\gamma=0.1)$. Both GD and
NGD are trained using a learning rate of $\eta=0.1$ and weight decay
of 0.01. For numerical stability in NGD, we apply a damping factor of
$\epsilon=10^{-3}$ to the FIM inversion.

All simulations were implemented in Python 3.13 using the \verb+NumPy+
2.1.3 and \verb+SciPy+ 1.15.2 libraries and were executed on a
standard workstation equipped with an Intel Core i9-9900K CPU and 64
GB of RAM. No GPU acceleration was used.

\section{The Geometry of Learning Trajectories}
\label{sec:results}

In this section, we present our main experimental results. We analyze
the learning trajectories of both GD and NGD on the Ridge of
Optimization, revealing a highly structured, geometry-driven learning
process.

\subsection{Two-Phase Learning via Pythagorean Decomposition}
\label{sec:two_phase}
To understand the temporal structure of the learning process, we
theoretically decompose the optimization trajectory. In information
geometry, the ``information gain'' (the reduction in KL-divergence)
during a small parameter update
$\boldsymbol{\delta}_t = \boldsymbol{\alpha}_{t+1} -
\boldsymbol{\alpha}_t$ can be approximated by a quadratic form of the
Fisher Information Matrix (FIM). We define the information gain at
step $t$ as:
\begin{equation}
    \Delta I_t \coloneqq \frac{1}{2} \boldsymbol{\delta}_t^\top G(\boldsymbol{\alpha}^*) \boldsymbol{\delta}_t,
    \label{eq:info_gain}
\end{equation}
where we use the FIM at the converged state,
$G(\boldsymbol{\alpha}^*)$, as a fixed global reference metric. This
choice, while a linearization of the true dynamics, allows us to
retrospectively analyze the trajectory's components within a
consistent coordinate system defined by the final solution's
geometry. A detailed justification for this approach and the
derivation of this quadratic form from KL-divergence are provided in
Appendix~\ref{app:pythagorean_decomp}.

Let
$G(\boldsymbol{\alpha}^*) = \sum_{k=1}^P \lambda_k \bm{v}_k
\bm{v}_k^\top$ be the eigendecomposition of the FIM.  We can
orthogonally project the information gain onto this eigenspace:
\begin{equation}
    \Delta I_t = \frac{1}{2} \lambda_1 (\bm{v}_1^\top \boldsymbol{\delta}_t)^2 + \frac{1}{2} \sum_{k=2}^P \lambda_k (\bm{v}_k^\top \boldsymbol{\delta}_t)^2.
    \label{eq:gain_decomposition}
\end{equation}
The first term represents the Principal Gain, capturing the
optimization progress along the dominant curvature direction
(associated with global attractor stability). The second term
represents the Tail Gain, reflecting the fine-tuning of the remaining
degrees of freedom (associated with memory capacity).

Based on the empirical property of Spectral Concentration on the Ridge
($\lambda_1 \gg \lambda_{k>1}$), we formulate the following
\textbf{heuristic proposition} regarding the initial GD learning
dynamics under a local quadratic approximation:

\begin{proposition}[Two-Phase Learning Dynamics]
  For GD operating on the Ridge of Optimization, where the FIM
  exhibits extreme spectral concentration
  ($\lambda_1 \gg \lambda_{k>1}$), the initial learning phase is
  primarily driven by the Principal Gain
  ($\Delta I_t \approx \frac{1}{2} \lambda_1 (\bm{v}_1^\top
  \boldsymbol{\delta}_t)^2$), while the subsequent phase is governed
  by the Tail Gain, commencing after the gradient component along
  $\bm{v}_1$ is substantially reduced
  ($\bm{v}_1^\top \nabla L \approx 0$).
\end{proposition}
This proposition serves as an interpretive model for the observed
dynamics, not a strict dynamical theorem.  For a rigorous dynamical
systems analysis of this spectral concentration and the interaction
between the maximum margin objective and the stability limit, we refer
the reader to our recent theoretical work~\cite{tamamori_eos_2026}.
It suggests that GD cannot effectively optimize all directions
simultaneously in such an anisotropic space; instead, the steep
geometry forces the trajectory to prioritize descent along the
principal curvature.

\begin{figure}[t]
  \centering  
  \begin{minipage}[b]{0.5\textwidth}
    \centering
    \includegraphics[width=\textwidth]{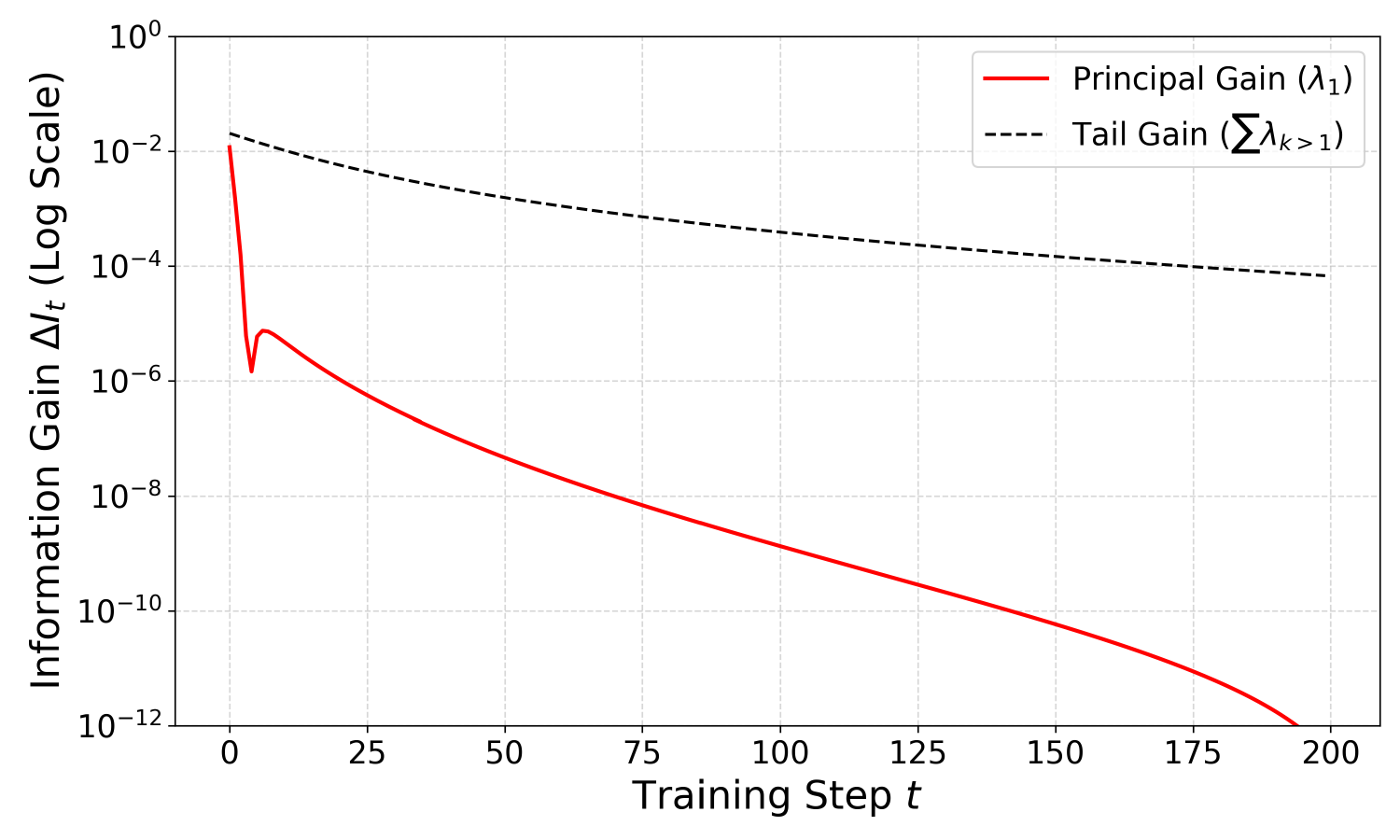}
    \centerline{\small (a) On the Ridge ($\gamma=0.02$)}
  \end{minipage}
  \hfill
  \begin{minipage}[b]{0.5\textwidth}
    \centering
    \includegraphics[width=\textwidth]{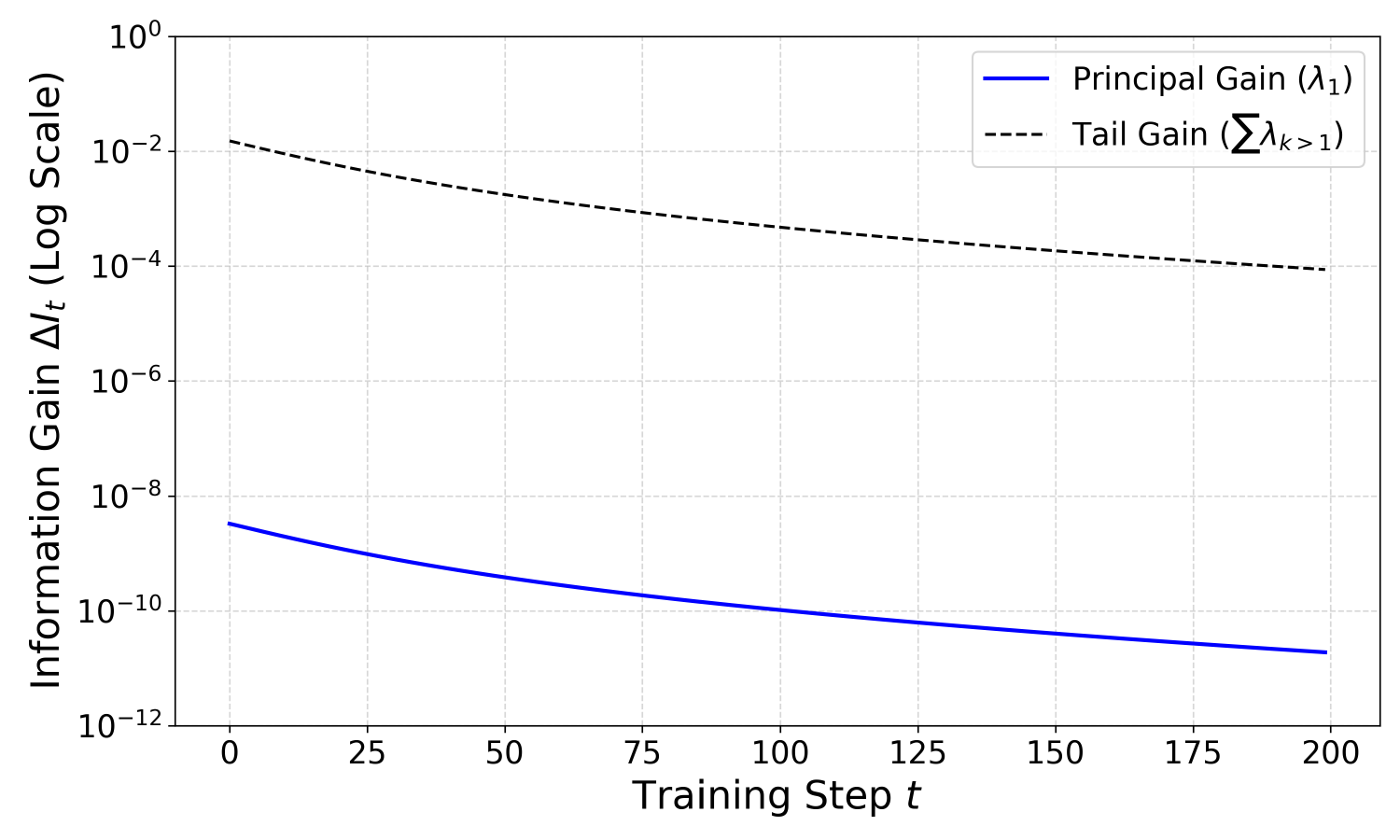}
    \centerline{\small (b) In the Local regime ($\gamma=0.1$)}
  \end{minipage}
  \caption{\textbf{Information Gain Decomposition.} The plot shows the
    information gain decomposed into Principal Gain (solid line) and
    Tail Gain (dashed line) for (a) the Ridge regime and (b) the Local
    regime.}
  \label{fig:info_gain}
\end{figure}

We empirically validated this proposition by tracking the decomposed
gains during training. Figure~\ref{fig:info_gain} compares this
decomposition for trajectories on the Ridge and in the Local
regime. As theoretically predicted, the learning process on the Ridge
(Fig.~\ref{fig:info_gain} (a)) clearly exhibits two distinct,
non-overlapping phases: an initial rapid spike in Principal Gain
($t < 25$) followed by a slow, extended decay in Tail Gain. In stark
contrast, the Local regime (Fig.~\ref{fig:info_gain} (b)), which lacks
spectral concentration ($\lambda_1 \approx \lambda_k$), shows no such
temporal separation, with both components decaying concurrently.

\subsection{Learning Trajectory on the Statistical Manifold}
\label{sec:trajectory}

\begin{figure}[h]
  \centering  
  \begin{minipage}[b]{0.5\textwidth}
    \centering
    \includegraphics[width=\textwidth]{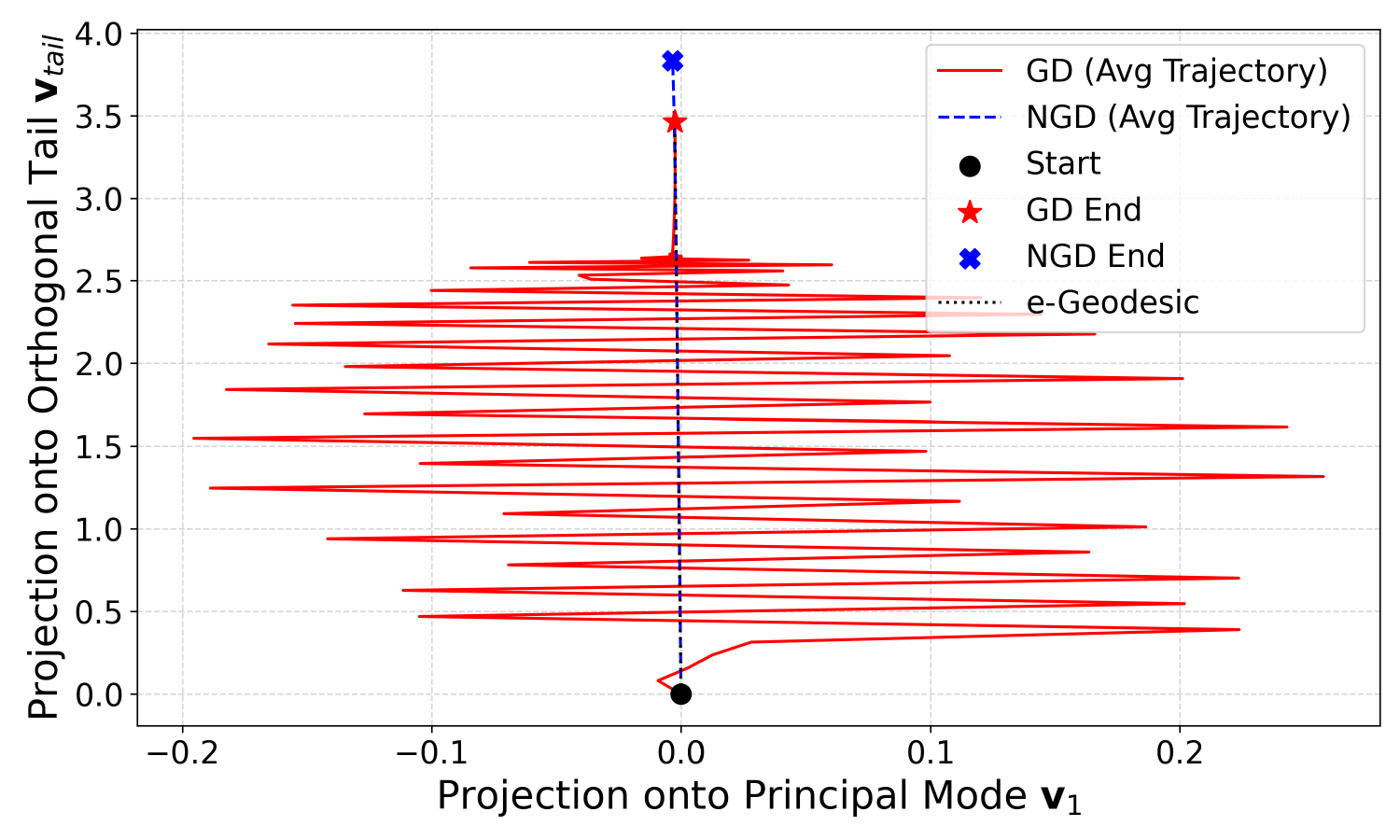}
    \centerline{\small (a) On the Ridge ($\gamma=0.02$)}
  \end{minipage}
  \hfill
  \begin{minipage}[b]{0.5\textwidth}
    \centering
    \includegraphics[width=\textwidth]{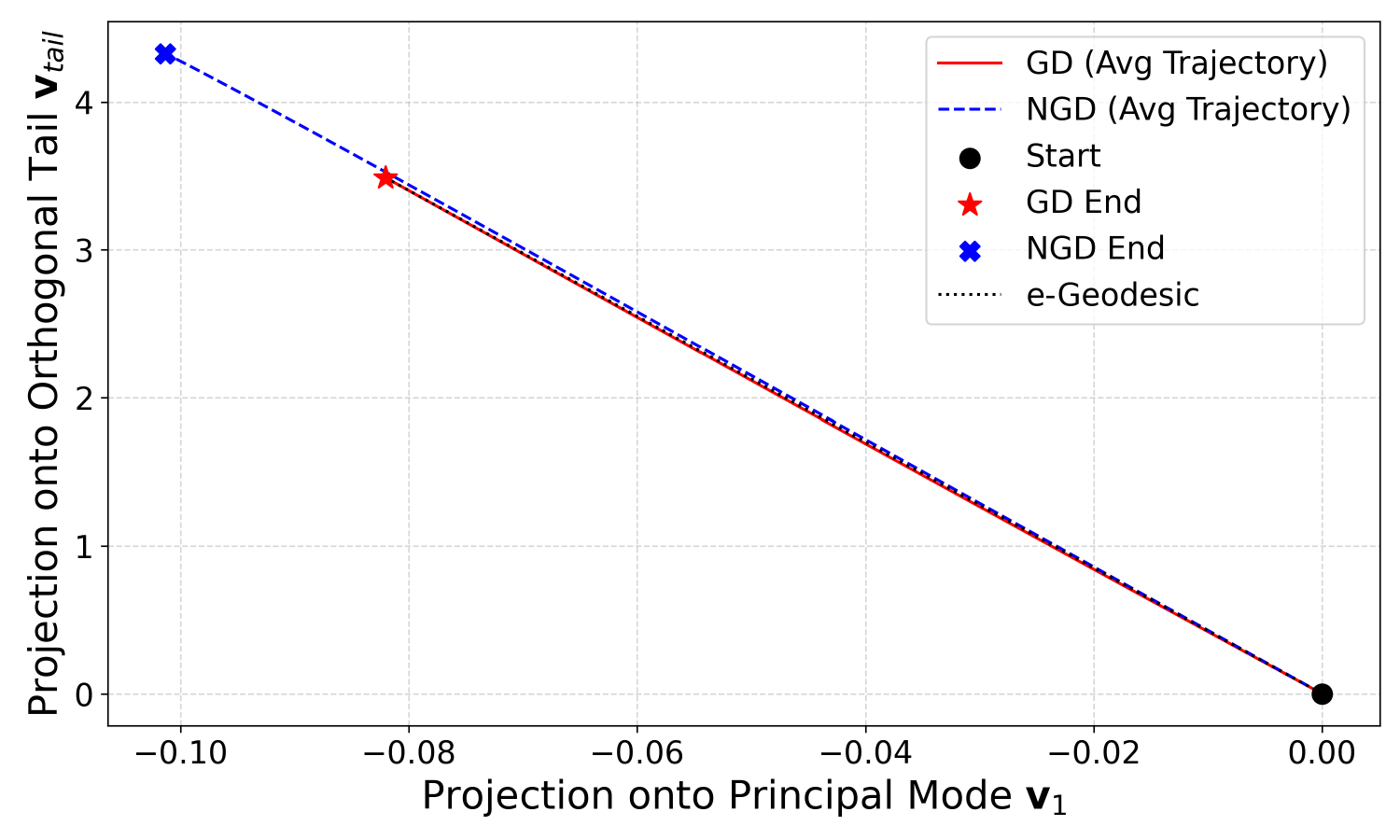}
    \centerline{\small (b) In the Local regime ($\gamma=0.1$)}
  \end{minipage}
  \caption{\textbf{Learning Trajectory Projection.} Comparison of GD
    and NGD trajectories projected onto the 2D eigenspace of the final
    FIM. \textbf{(a) On the Ridge ($\gamma=0.02$):} The GD trajectory
    follows a highly oscillatory, non-geodesic path, strongly
    deviating from the $e$-geodesic, while the NGD trajectory (blue
    dashed line) almost perfectly overlaps with the ideal $e$-geodesic
    (black dotted line).  \textbf{(b) In the Local Regime
      ($\gamma=0.1$):} Both trajectories are nearly straight,
    indicating a flat learning landscape.}
  \label{fig:trajectory_projection}
\end{figure}

To visualize how the optimization dynamics navigate the highly skewed
geometry of the Ridge, we projected the learning trajectories onto a
2D plane. This plane is spanned by the principal eigenvector
$\bm{v}_1$ of the final Fisher Information Matrix (FIM) and an
orthogonal direction vector $\bm{v}_{\text{tail}}$. The vector
$\bm{v}_{\text{tail}}$ is defined as the normalized projection of the
final converged state $\boldsymbol{\alpha}^*$ onto the subspace
orthogonal to $\bm{v}_1$. We compared the trajectory of standard
Gradient Descent (GD) with that of Natural Gradient Descent (NGD).

Figure~\ref{fig:trajectory_projection} displays the averaged
trajectories across all neurons. In the Local regime ($\gamma=0.1$,
Fig.~\ref{fig:trajectory_projection} (b)), where the FIM spectrum is
relatively flat, both GD and NGD follow nearly identical, direct
paths toward the optimal solution. This straight path corresponds
closely to the $e$-geodesic in the parameter space.
However, a stark contrast emerges on the Ridge of Optimization
($\gamma=0.02$, Fig.~\ref{fig:trajectory_projection} (a)). The NGD
trajectory (blue dashed line) continues to follow the ideal $e$-geodesic
almost perfectly. This indicates that by explicitly correcting for the
manifold's curvature using the inverse FIM, NGD maintains a direct and
efficient path to the minimum. Conversely, the GD trajectory (red
solid line) exhibits severe oscillations along the principal direction
$\bm{v}_1$.

This oscillatory behavior is a direct consequence of the extreme
spectral concentration on the Ridge. The massive curvature associated
with $\lambda_1$ causes the standard Euclidean gradient step to
overshoot the narrow valley of the loss landscape, a phenomenon
closely related to the Edge of Stability observed in deep neural
networks~\cite{Cohen2021}.  As detailed in~\cite{tamamori_eos_2026},
this overshooting triggers a transient self-stabilizing feedback loop
governed by the logistic variance, which temporarily pins the
principal curvature near the stability limit $2/\eta$.  Despite these
violent oscillations in the principal direction, the GD trajectory
gradually progresses along the orthogonal tail subspace to eventually
reach the vicinity of the optimal solution. This visualization
confirms that the geometry of the Ridge imposes severe constraints on
standard gradient methods, forcing them into highly inefficient,
oscillatory paths. To ensure that these observations are not an
artifact of a specific hyperparameter choice, we conducted additional
experiments across a wider range of storage loads and kernel
localities (see Appendix~\ref{app:generality}). These supplementary
results confirm that the severe oscillatory behavior is a robust
signature of GD specifically on the Ridge under high memory
congestion.

\begin{figure}[t]
  \centering
  \includegraphics[width=\columnwidth]{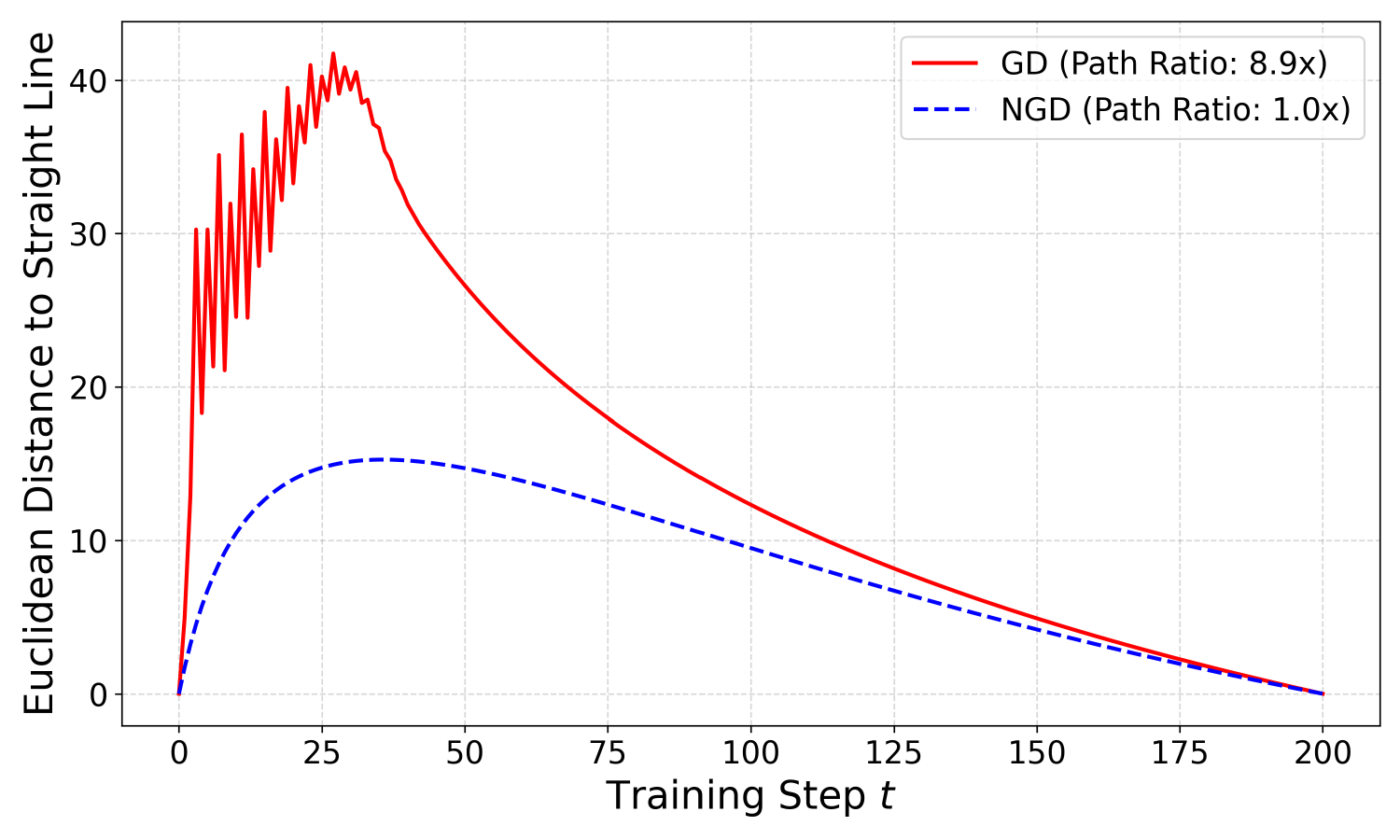}
  \caption{\textbf{Quantitative Analysis of Learning Trajectories.}
    The plot shows the Euclidean distance from the learning
    trajectory to the direct straight-line path connecting the
    initial and final parameter states on the Ridge ($\gamma=0.02$).
  }
  \label{fig:geometric_metrics}
\end{figure}

To rigorously quantify these geometric differences beyond 2D
projections, we measured two metrics. First, we computed the Euclidean
distance from the learning trajectory $\boldsymbol{\alpha}(t)$ to the
direct straight line connecting the initial state
$\boldsymbol{\alpha}_{0}$ and the final state
$\boldsymbol{\alpha}^*$. Second, we calculated the Path Ratio, defined
as the cumulative path length of the trajectory divided by the direct
Euclidean distance
$\|\boldsymbol{\alpha}^* - \boldsymbol{\alpha}_{0}\|$.
Figure~\ref{fig:geometric_metrics} plots the Euclidean distance to
this straight line at each training step.

The results provide a striking quantitative contrast. The GD
trajectory (red solid line) deviates massively from the direct path,
with severe oscillations characterizing its initial phase. This
overshooting is a direct manifestation of the Edge of Stability
phenomenon, where the extreme curvature on the Ridge destabilizes
standard gradient descent. Consequently, the cumulative path length of
GD is nearly 9 times longer than the direct distance (Path Ratio
$\approx 8.9\times$), confirming its highly inefficient, non-geodesic
navigation.

In contrast, the NGD trajectory (blue dashed line) is remarkably
smooth and stable. While it exhibits a slight, smooth deviation from
the Euclidean straight line, which is a natural consequence of
following the true curved geodesic on the non-Euclidean statistical
manifold, its total path length is nearly optimal (Path Ratio
$\approx 1.0\times$). This quantitative analysis confirms that the
specific geometry of the Ridge imposes severe constraints on standard
gradient methods, while information-geometric optimization (NGD)
effectively mitigates these instabilities by aligning perfectly with
the intrinsic curvature of the memory landscape.

\section{Geometric Alignment and Generalization on the Ridge}
\label{sec:optimality}
The trajectory analysis in Section~\ref{sec:trajectory} demonstrated
that GD struggles with the extreme curvature of the Ridge, whereas NGD
navigates it smoothly. We now evaluate the practical implications of
these geometric differences by comparing the convergence speed and
generalization performance of the two algorithms. Our goal here is not
to propose NGD as a computationally competitive alternative to modern
adaptive optimizers (e.g., Adam~\cite{Kingma2015}), but rather to use
it as a theoretical baseline that perfectly aligns with the intrinsic
geometry of the statistical manifold.  We evaluated the models on a
held-out validation dataset (comprising 20\% of the generated
patterns) during training.

\subsection{Overcoming Instability at the Edge of Stability}
The Ridge of Optimization is defined by an extremely large maximal
eigenvalue of the FIM, $\lambda_{\max}(G)$, which can grow by orders
of magnitude with the storage load. This implies that the statistical
manifold is highly anisotropic, forming a steep potential valley. In
optimization theory, such a condition is closely associated with the
Edge of Stability~\cite{Cohen2021}, where the large curvature
$\lambda_1$ exceeds the stability limit $2/\eta$ of standard GD,
causing the learning dynamics to bounce violently between the walls of
the valley.

We investigated this instability by comparing the learning curves of
GD and NGD. While it is theoretically expected that NGD, which
pre-conditions the gradient with the inverse FIM $(G^{-1}$, should
converge faster in mildly curved spaces, its behavior on the Ridge is
non-trivial. Because the FIM is nearly singular
($\lambda_1 \gg \lambda_{k>1} \approx 0$), computing $G^{-1}$ is
numerically ill-posed, and one might expect NGD to fail or require
prohibitively small learning rates in this extreme regime.

\begin{figure}[t]
  \centering  
  \begin{minipage}[b]{0.5\textwidth}
    \centering
    \includegraphics[width=\textwidth]{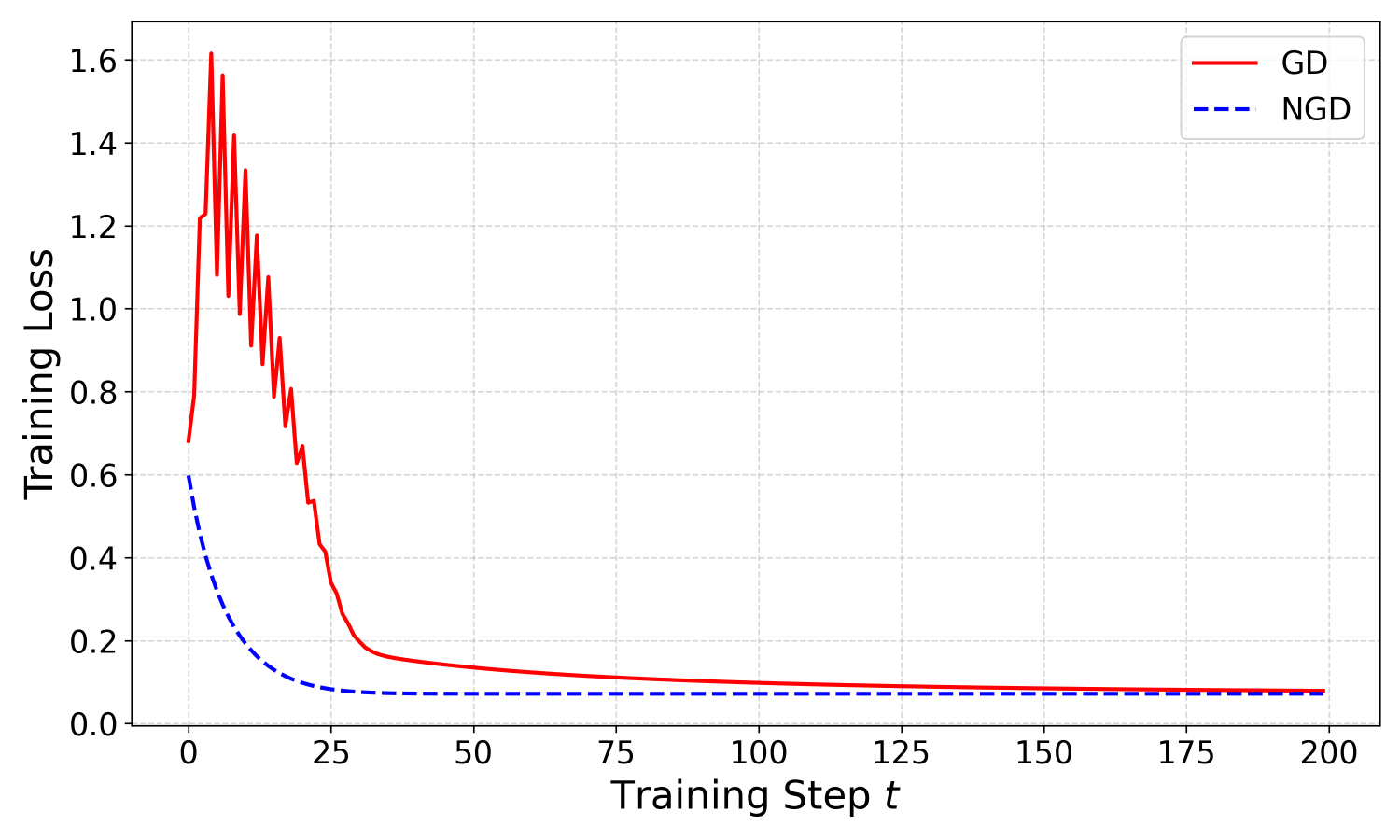}
    \centerline{\small (a) Training Loss}
  \end{minipage}
  \hfill
  \begin{minipage}[b]{0.5\textwidth}
    \centering
    \includegraphics[width=\textwidth]{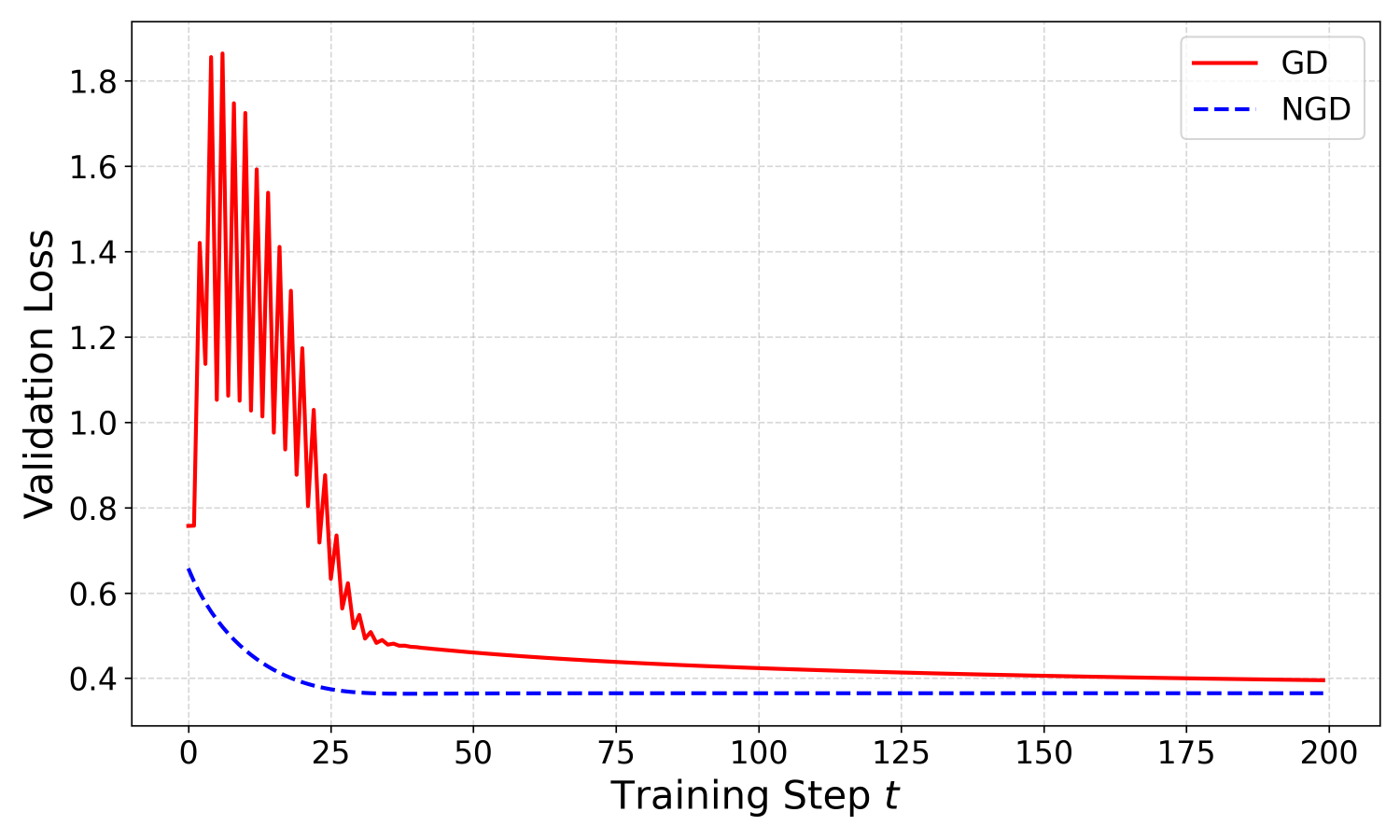}
    \centerline{\small (b) Validation Loss}
  \end{minipage}
  \caption{\textbf{Learning Curves for GD vs. NGD.} Comparison of (a)
    Training Loss and (b) Validation Loss on the Ridge
    ($\gamma=0.02$). NGD converges faster and achieves a lower final
    validation loss, demonstrating superior optimization and
    generalization performance.}
  \label{fig:learning_curves}
\end{figure}

However, as shown in Fig.~\ref{fig:learning_curves} (a), the GD
trajectory (red line) exhibits the expected large oscillations in the
initial phase, a clear signature of overshooting the principal
curvature. In stark contrast, the NGD trajectory (blue dashed line),
stabilized with a small damping factor, shows a smooth, monotonic
decrease. By effectively ``flattening'' the highly skewed landscape
via inverse FIM preconditioning, NGD effectively mitigates the Edge of
Stability phenomenon. This demonstrates that the extreme geometric
structure of the Ridge (Spectral Concentration) is not merely an
obstacle, but a landscape well-aligned with the mechanics of
information-geometric optimization. The fact that NGD succeeds so
robustly in a regime where GD severely oscillates highlights the
fundamental role of intrinsic curvature in high-capacity memory
formation.

\subsection{Generalization Performance: GD vs. NGD}
Faster convergence does not necessarily imply a better solution. A
crucial question is whether the oscillatory dynamics of GD confer any
advantage in terms of generalization, for instance by helping to find
``flatter'' minima. To test this, we evaluated the generalization
performance by monitoring the loss on a held-out validation set during
training.

Figure~\ref{fig:learning_curves} (b) plots the validation loss for
both methods. The results clearly show that NGD not only converges
faster but also achieves a consistently lower final validation loss
compared to GD. This indicates that the smooth, geodesic path taken
by NGD leads to a solution with superior generalization capabilities
in this specific setting. The oscillations inherent to GD on the
Ridge do not appear to confer any regularization benefit in this
context; rather, they are a suboptimal consequence of navigating a
curved statistical manifold with a Euclidean metric.

These findings provide strong evidence that the geometric structure of
high-capacity associative memory is well-aligned with natural gradient
methods, suggesting that information-geometric optimization provides a
theoretically sound approach for learning on the
Ridge.

\section{Discussion}
\label{sec:discussion}

In this study, we have provided a detailed geometric analysis of the
learning dynamics in high-capacity KLR Hopfield networks. By comparing
the trajectories of GD and NGD, we have uncovered a highly
structured, geometry-driven optimization process. Here, we discuss the
broader implications of these findings.

\subsection{GD's Oscillatory Dynamics as a Heuristic for Feature Learning}
Our results offer a new interpretation of the learning path taken by
standard GD. The observed trajectory
(Fig.~\ref{fig:trajectory_projection}), characterized by an initial
rapid phase of severe oscillations along the principal curvature
direction followed by a slow fine-tuning phase, can be seen as a
sophisticated, albeit implicit, strategy for hierarchical feature
learning. GD, without any explicit knowledge of the manifold's
geometry, effectively prioritizes learning the most dominant
structural features (corresponding to $\lambda_1$) before refining the
details. This two-phase process, empirically observed in our
information gain decomposition (Fig.~\ref{fig:info_gain}), provides a
geometric explanation for how simple optimizers can navigate complex,
highly anisotropic loss landscapes, despite the inherent instabilities
at the Edge of Stability.

\subsection{The Optimality of Natural Gradient on the Ridge}
A key finding of this work is the superior performance of NGD on the
Ridge of Optimization (Fig.~\ref{fig:learning_curves}). The Ridge is a
regime of extreme curvature, where the FIM is nearly singular. For GD,
this sharp geometry leads to instability and violent oscillations, a
manifestation of the Edge of Stability phenomenon. NGD, however,
thrives in this environment. By explicitly inverting the FIM, it
``flattens'' the landscape and follows the true geodesic path,
achieving faster convergence and better generalization. This suggests
that the geometric structures emerging during high-capacity learning
(i.e., Spectral Concentration) are not mere obstacles, but rather are
optimally matched to information-geometric optimization methods.

\subsection{Rethinking the Role of ``Flat Minima''}
A popular hypothesis in deep learning states that GD preferentially
finds ``flat minima'', which are associated with better
generalization. Our results present a more nuanced picture. The
solution on the Ridge is, by definition, extremely ``sharp'' in the
dominant spectral direction. GD struggles in this sharp valley, while
NGD finds the minimum efficiently. Yet, the NGD solution exhibits
superior generalization. This suggests that for certain structured
problems like associative memory, the relevant geometric property may
not be the flatness of the minimum itself, but rather the alignment of
the learning algorithm with the intrinsic curvature of the data
manifold. NGD achieves this alignment by construction, leading to a
high-quality solution in a sharp but well-structured energy basin.

\subsection{Limitations and Future Work}
\label{sec:limitations}

While this study provides a detailed geometric picture of learning
dynamics in a specific, highly structured regime, we acknowledge
several limitations that open avenues for future research.

First, our analysis is primarily focused on the KLR-trained Hopfield
network with an RBF kernel, trained on uncorrelated random
patterns. While this idealized setting is crucial for isolating the
fundamental geometric and spectral mechanisms, it remains an open
question how these dynamics translate to networks with different
kernel functions (e.g., polynomial) or to tasks involving structured,
real-world data such as natural images or language. Investigating how
data correlations alter the geometry of the Ridge and the resulting
learning trajectories is a key next step.

Second, our comparison was limited to vanilla GD and exact NGD. While
NGD served as an ideal theoretical baseline to reveal the underlying
geometry, it remains a first-order manifold optimization method, which
may exhibit slow convergence near the optimum. Recent advancements in
information geometry have proposed second-order methods, such as the
Dual Riemannian Newton Method~\cite{Zhou2025}, that leverage dual
affine connections to achieve local quadratic convergence. Applying
such advanced methods to the extreme-curvature environment of the
Ridge presents a highly promising avenue for achieving even faster and
more stable memory formation.

Third, a related practical limitation is the computational cost. A
significant barrier to the application of both exact NGD and
Newton-type methods is the cost of forming and inverting the FIM or
Hessian, which scales as $O(P^3)$ per update. Fortunately, the highly
skewed spectral structure of the Ridge suggests several paths
forward. One approach is to utilize scalable, low-rank approximation
methods, such as K-FAC~\cite{Martens2015}, which are particularly
well-suited to landscapes dominated by a few large
eigenvalues. Another promising direction is to employ adaptive natural
gradient algorithms that recursively update the inverse FIM without
requiring full matrix inversion at each step~\cite{Park2000,
  Amari2006}, thereby potentially reducing the per-update complexity
to $O(P^{2})$. Future work should investigate whether these more
practical adaptive optimizers can replicate the stable, geodesic-like
convergence of the exact NGD observed on the Ridge.

Finally, our study focused on the learning process itself. Extending
this geometric framework to analyze the robustness of the final
solution against different types of perturbations, such as adversarial
attacks or data drift, would be a valuable direction for future work.

\section{Conclusion}
\label{sec:conclusion}

In this work, we have presented a comprehensive geometric analysis of
the learning dynamics in high-capacity KLR-trained Hopfield
networks. By comparing the trajectories of standard GD and
information-geometric NGD, we have moved beyond a static analysis of
attractors to a dynamic understanding of how optimal memory
representations are formed.

Our analysis revealed that learning on the Ridge of Optimization is a
highly structured, two-phase process. The GD trajectory, guided
implicitly by the manifold's extreme curvature, follows a non-geodesic
path characterized by severe initial oscillations, prioritizing the
acquisition of global stability before fine-tuning for capacity. Most
importantly, we demonstrated that the sharp curvature of the Ridge,
while presenting a formidable challenge for GD, provides an ideal
landscape for NGD. By explicitly correcting for the geometry, NGD
achieves faster, more stable convergence and finds a solution with
superior generalization performance.

These findings clarify the geometric principles underlying
self-organization in kernel associative memory and provide strong
evidence for the theoretical optimality of information-geometric
optimization methods in regimes of high spectral concentration. This
work opens new avenues for developing more efficient learning
algorithms for high-capacity memory systems and deepens our
understanding of the interplay between learning dynamics and the
emergent geometry of neural representations.

\funding
Not applicable.

\conflictsofinterest
The author declares no competing interests.

\authorcontribution
The sole author contributed to the present work.

\aitools
The author utilized AI language models (GPT-5.6 Luna and Gemini 3.1
Pro) to assist with brainstorming, preliminary mathematical
derivations, and English proofreading during the preparation of this
manuscript. The author thoroughly reviewed, verified, and edited all
AI-generated content and takes full responsibility for the final
contents of the publication.

\appendix
\section{Information Gain Decomposition}
\label{app:pythagorean_decomp}

This appendix provides a brief derivation of the information gain
decomposition used in Section~\ref{sec:two_phase}.

\subsection{Information Gain as KL-Divergence}
The progress of learning in one step can be measured by the reduction
in the KL-divergence between the target data distribution and the
model distribution. For Maximum Likelihood Estimation, this is
approximately equal to the KL-divergence between the model at step $t$
and step $t+1$:
\begin{equation}
    \Delta I_t \approx D_{KL}(P(\boldsymbol{\alpha}_{t+1}) || P(\boldsymbol{\alpha}_t)).
\end{equation}
For an infinitesimal parameter change
$\boldsymbol{\delta}_t = \boldsymbol{\alpha}_{t+1} -
\boldsymbol{\alpha}_t$, this KL-divergence can be approximated to
second order by the quadratic form of the FIM at the current point,
$G(\boldsymbol{\alpha}_t)$:
\begin{equation}
    \Delta I_t \approx \frac{1}{2} \boldsymbol{\delta}_t^\top G(\boldsymbol{\alpha}_t) \boldsymbol{\delta}_t.
\end{equation}

\subsection{Choice of a Fixed Reference Metric}
To analyze the global structure of the learning trajectory, it is
advantageous to use a fixed coordinate system rather than the evolving
instantaneous metric $G(\boldsymbol{\alpha}_t)$. We therefore choose
the FIM at the converged solution, $G(\boldsymbol{\alpha}^*)$, as a
global reference metric. This approach linearizes the manifold
geometry around the final solution and allows us to decompose the
entire trajectory into components that are globally orthogonal with
respect to this final geometry.  Thus, for our analysis, we define the
information gain as the projection onto this final metric:
\begin{equation}
  \Delta I_t \coloneqq \frac{1}{2} \boldsymbol{\delta}_t^\top G(\boldsymbol{\alpha}^*) \boldsymbol{\delta}_t.
\end{equation}
This allows us to track how much of the ``effort'' in each update step
contributes to forming the principal and tail components of the final
geometric structure. This simplification, while not capturing the full
non-linear dynamics, effectively reveals the hierarchical, two-phase
nature of the learning process on the Ridge.

\section{Generality of the Oscillatory Dynamics}
\label{app:generality}

\begin{figure*}[ht]
    \centering
    \begin{minipage}{0.49\textwidth}
        \centering
        \includegraphics[width=\linewidth]{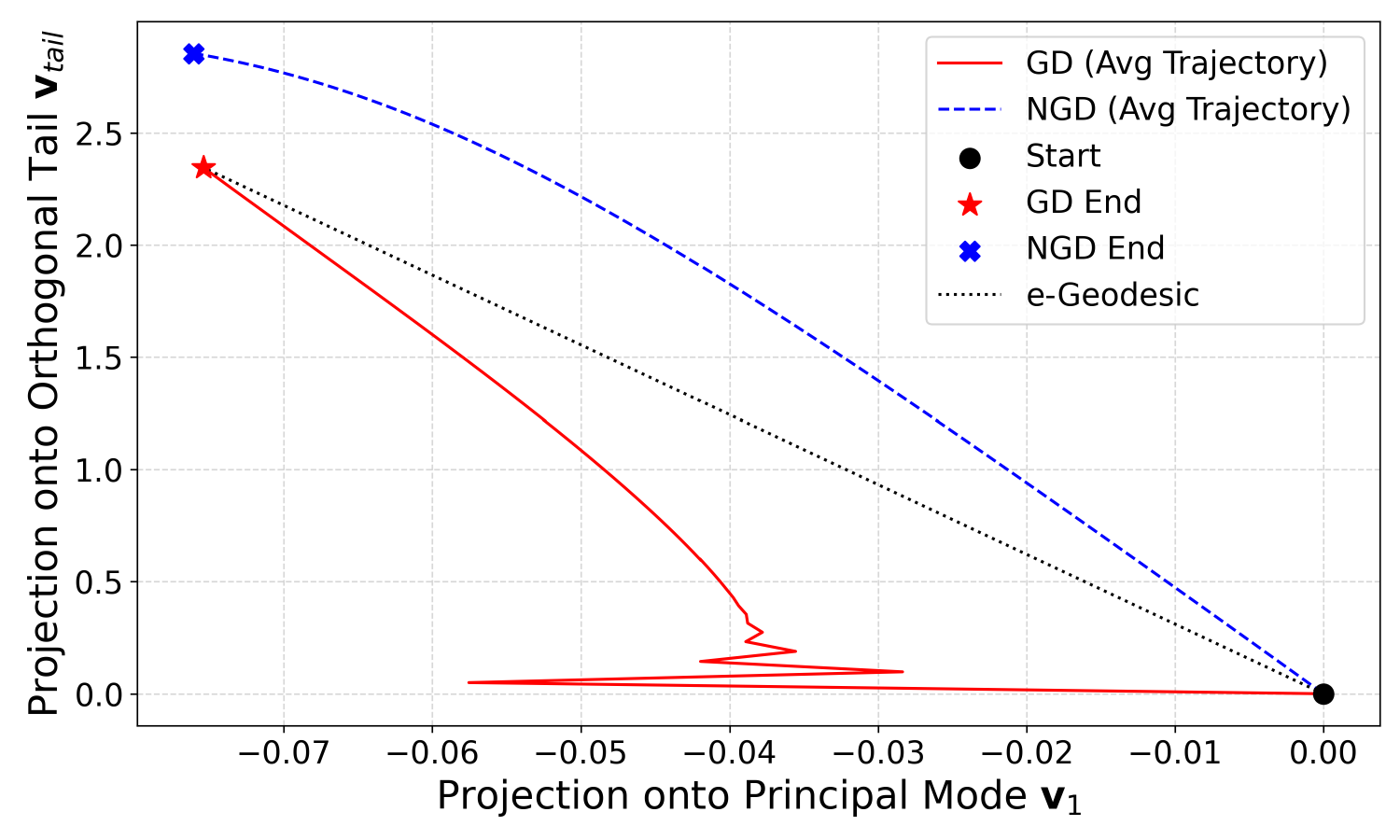}
        {\small (a) Low Load on Ridge ($P/N=1.0, \gamma=0.02$)}
    \end{minipage}\hfill
    \begin{minipage}{0.49\textwidth}
        \centering
        \includegraphics[width=\linewidth]{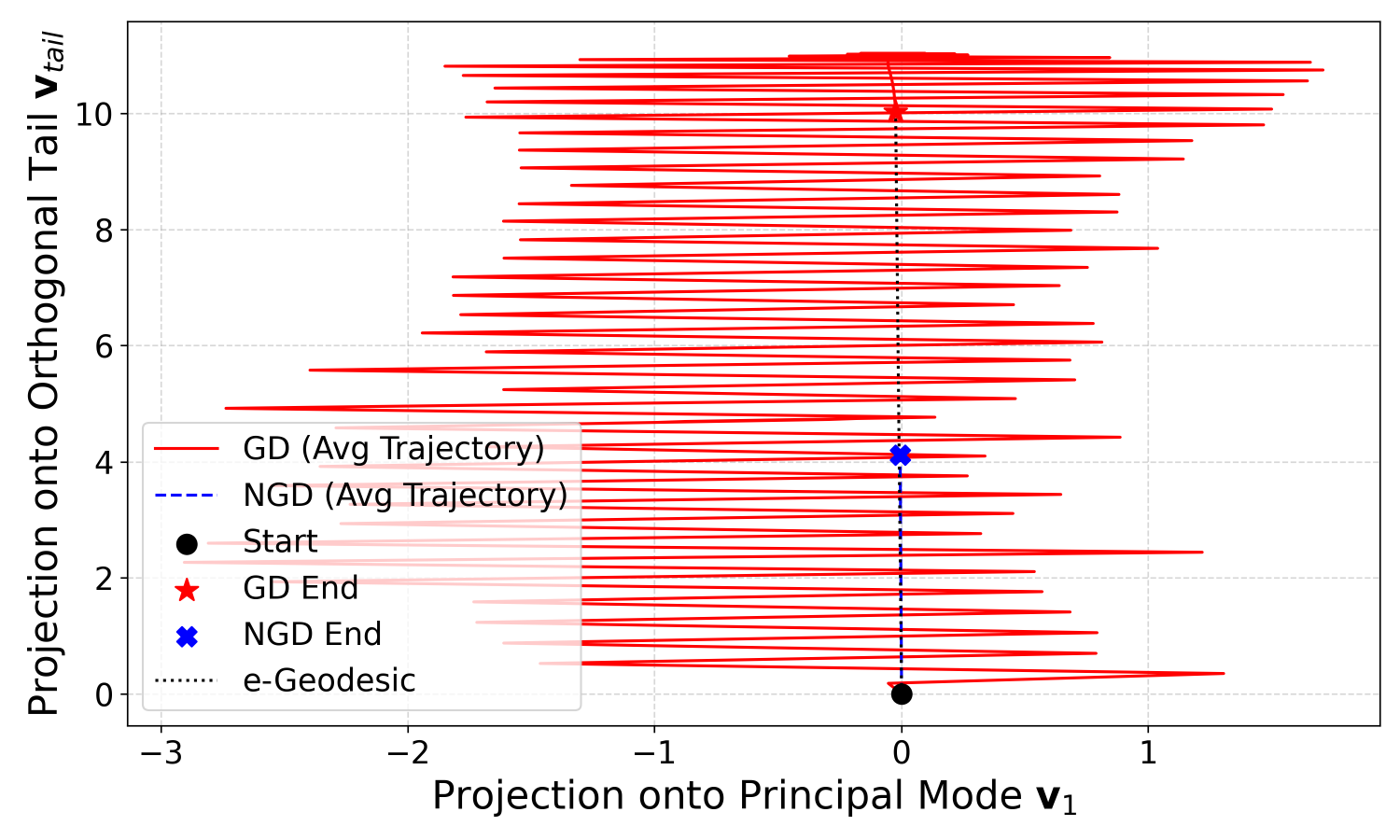}
        {\small (b) High Load on Ridge ($P/N=4.0, \gamma=0.02$)}
    \end{minipage}
    
    \vspace{0.5cm}
    
    \begin{minipage}{0.49\textwidth}
        \centering
        \includegraphics[width=\linewidth]{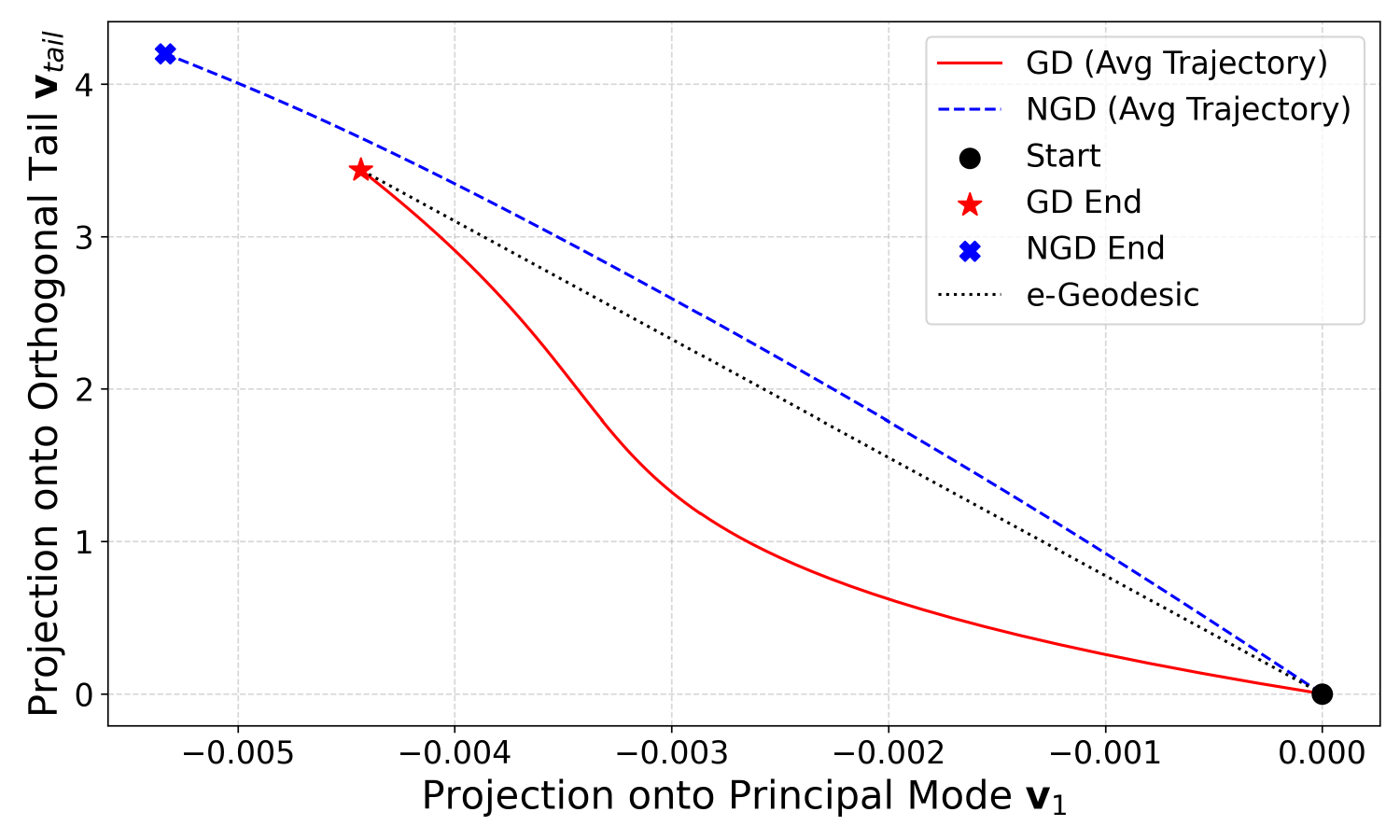}
        {\small (c) Medium Load in an Intermediate Regime ($P/N=2.0, \gamma=0.05$)}
    \end{minipage}\hfill
    \begin{minipage}{0.49\textwidth}
        \centering
        \includegraphics[width=\linewidth]{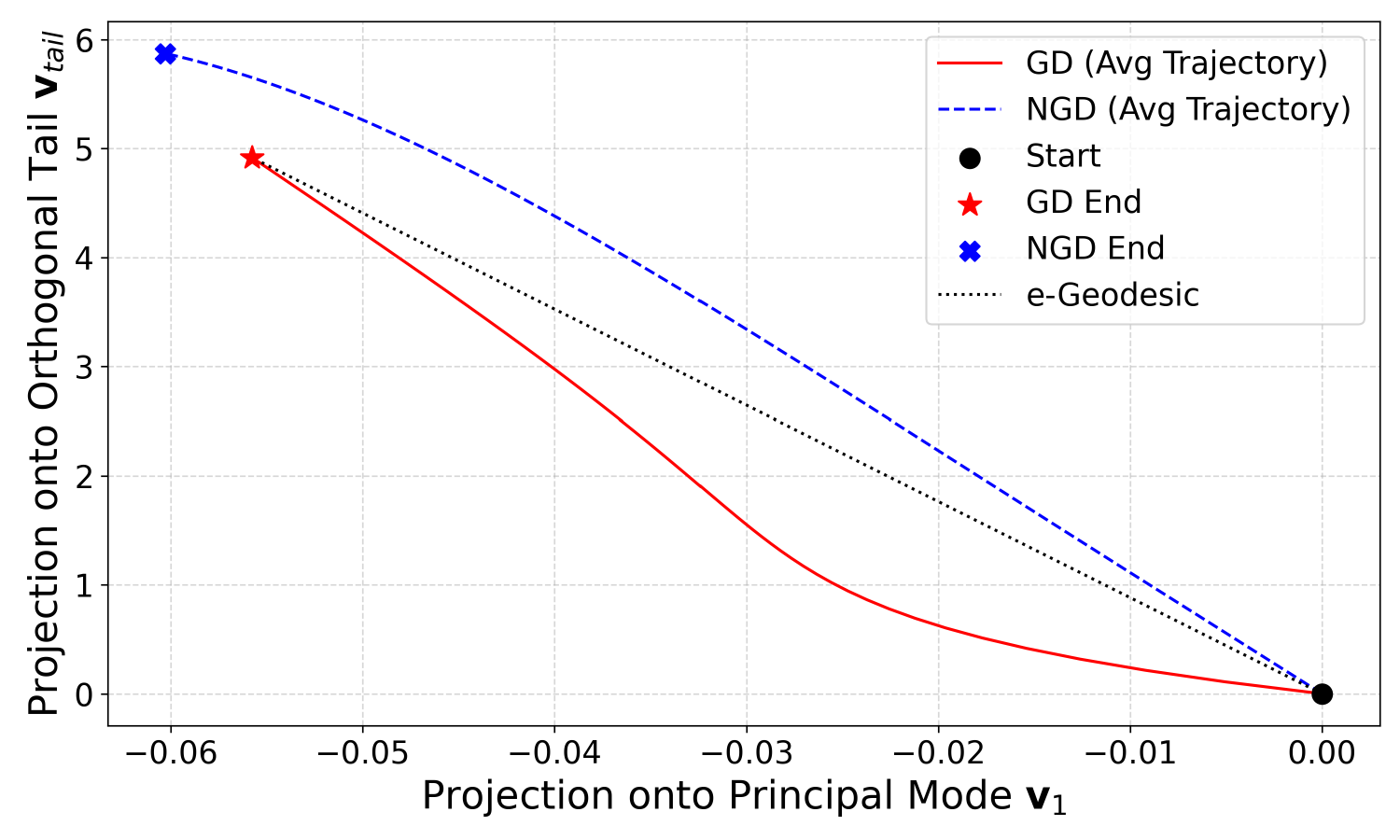}
        {\small (d) High Load in an Intermediate Regime ($P/N=4.0, \gamma=0.05$)}
    \end{minipage}
    
    \caption{\textbf{Generality of Learning Trajectories.} 2D
      projections of GD (red) and NGD (blue dashed) trajectories
      across varying hyperparameter regimes. The top row shows the
      effect of storage load on the Ridge, highlighting the emergence
      of severe oscillations at high loads (b). The bottom row shows
      the effect of relaxing the kernel locality ($\gamma=0.05$),
      where GD follows a smoother parabolic path without
      oscillations, regardless of the load. 
    }
    \label{fig:generality_trajectories}
\end{figure*}

To address potential concerns regarding the generality of the observed
oscillatory behavior of GD, we conducted additional experiments
across different hyperparameter regimes. Specifically, we varied the
storage load $P/N$ and the kernel locality parameter $\gamma$ to
investigate how the learning trajectories adapt to the changing
curvature of the statistical manifold.

Figure~\ref{fig:generality_trajectories} presents the 2D projected
trajectories arranged in a $2 \times 2$ matrix to compare the effects
of load and locality.  First, we examined the effect of the storage
load on the Ridge ($\gamma=0.02$, top row). At a low load of $P/N=1.0$
(Fig.~\ref{fig:generality_trajectories} (a)), the principal curvature
is relatively small. Consequently, GD does not exhibit severe
oscillations, but rather follows a smooth, curved path toward the
solution. Interestingly, the NGD trajectory slightly deviates from the
Euclidean straight line ($e$-geodesic), reflecting the non-trivial
Riemannian curvature of the manifold when it is not entirely dominated
by a single massive eigenvalue. In stark contrast, at a very high load
of $P/N=4.0$ (Fig.~\ref{fig:generality_trajectories} (b)), the extreme
spectral concentration exacerbates the Edge of Stability phenomenon,
causing GD to exhibit violent, prolonged oscillations.

Next, we investigated an intermediate ``Local'' regime by relaxing the
kernel locality to $\gamma=0.05$ (bottom row). Here, the curvature is
less extreme than on the optimal Ridge. For both moderate load
($P/N=2.0$, Fig.~\ref{fig:generality_trajectories} (c)) and high load
($P/N=4.0$, Fig.~\ref{fig:generality_trajectories} (d), the GD
trajectory loses its severe oscillatory nature and instead forms a
smooth parabolic arc, demonstrating a continuous transition to a more
direct descent.

These supplementary results confirm that the highly oscillatory,
non-geodesic path is a robust and characteristic feature of GD
specifically on the Ridge of Optimization under high-load conditions,
while NGD consistently exploits the intrinsic geometry across all
tested regimes.


\begin{thebibliography}{99}

\bibitem{tamamori2025}
  A.~Tamamori, ``Kernel logistic regression learning for high-capacity
  hopfield networks,'' \textit{IEICE Trans. Inf. \& Syst.}, vol.~E109-E, no.~2,
  pp.~293--297, February 2026.
  DOI:\href{https://doi.org/10.1587/transinf.2025EDL8027}{10.1587/transinf.2025EDL8027}

\bibitem{tamamori_nolta_a}
  A.~Tamamori, ``Quantitative attractor analysis of high-capacity
  kernel hopfield networks,'' \textit{NOLTA}, vol.~E17-N,
  no.~3, pp.~770--787, July 2026.
  DOI:\href{https://doi.org/10.1587/nolta.17.770}{10.1587/nolta.17.770}

\bibitem{tamamori_nolta_b}
  A.~Tamamori, ``Self-organization and spectral mechanism of attractor
  landscapes in high-capacity kernel hopfield networks,''
  \textit{NOLTA}, vol.~E17-N, no.~3, pp.~788--804, July 2026.
  DOI:\href{https://doi.org/10.1587/nolta.17.788}{10.1587/nolta.17.788}

\bibitem{Amari1998}
  S.~Amari, ``Natural gradient works efficiently in learning,'' \textit{Neural
  Computation}, vol.~10, no.~2, pp.~251--276, February 1998.
  DOI:\href{https://doi.org/10.1162/089976698300017746}{10.1162/089976698300017746}

\bibitem{Amari2016}
  S.~Amari, \textit{Information geometry and its applications}, Springer, February 2016.
  DOI:\href{https://doi.org/10.1007/978-4-431-55978-8}{10.1007/978-4-431-55978-8}

\bibitem{Raskutti2015}
  G.~Raskutti and S.~Mukherjee, ``The information geometry of mirror
  descent,'' \textit{IEEE Transactions on Information Theory}, vol.~61, no.~3,
  pp.~1451--1457, March 2015.
  DOI:\href{https://doi.org/10.1109/TIT.2015.2388583}{10.1109/TIT.2015.2388583}

\bibitem{Jacot2018}
  A.~Jacot, F.~Gabriel, and C.~Hongler, ``Neural tangent kernel:
  convergence and generalization in neural networks,'' Proc. NIPS'18,
  pp.~8580--8589, December 2018.

\bibitem{Keskar2017}  
  N.S.~Keskar, D.~Mudigere, J.~Nocedal, M.~Smelyanskiy, and P.T.P~Tang,
  ``On large batch training for deep learning: generalization gap and
  sharp minima,'' Proc. ICLR'17, April 2017.

\bibitem{Sagun2018}
  L.~Sagun, U.~Evci, V.U.~G\"{u}ney, Y.~Dauphin, L.~Bottou,
  ``Empirical analysis of the hessian of over-parametrized neural
  networks,'' arXiv preprint arXiv:1706.04454, June 2017.
  DOI:\href{https://doi.org/10.48550/arXiv.1706.04454}{10.48550/arXiv.1706.04454}
  
\bibitem{Cohen2021}
  J.M.~Cohen, S.~Kaur, Y.~Li, J.Z.~Kolter, and A.~Talwalkar, ``Gradient
  descent on neural networks typically occurs at the edge of
  stability,'' Proc. ICLR'21, May 2021.

\bibitem{tamamori_eos_2026}
  A.~Tamamori, ``Information Geometric Self-Organization at the Edge
  of Stability in High-Capacity Kernel Associative Memories,'' arXiv
  preprint arXiv:2609.16827, September 2026.
  DOI:\href{https://doi.org/10.48550/arXiv.2609.16827}{10.48550/arXiv.2609.16827}

\bibitem{Kingma2015}  
  D.~Kingma and J.~Ba, ``Adam: a method for stochastic optimization,''
  Proc. ICLR'15, May 2015.

\bibitem{Zhou2025}
  D.~Zhou, K.~Yano and M.~Sugiyama, ``Dual riemannian newton method on
  statistical manifolds,'' arXiv preprint arXiv:2511.11318, November
  2025.
  DOI:\href{https://doi.org/10.48550/arXiv.2511.11318}{10.48550/arXiv.2511.11318}

\bibitem{Martens2015}
  J.~Martens and R.~Grosse, ``Optimizing neural networks with
  kronecker-factored approximate curvature,'' Proc. ICML'15,
  vol.~37. pp.~2408--2417, July 2015.

\bibitem{Park2000}
  H.~Park, S.~Amari, and K.~Fukumizu, ``Adaptive natural gradient
  learning algorithms for various stochastic models,'' \textit{Neural Networks},
  vol.~13, no.~7, pp.~755--764, September 2000.
  DOI:\href{https://doi.org/10.1016/S0893-6080(00)00051-4}{10.1016/S0893-6080(00)00051-4}
  
\bibitem{Amari2006}
  S.~Amari, H.~Park, and T~Ozeki, ``Singularities affect dynamics of
  learning in neuromanifolds,'' \textit{Neural Computation}, vol.~18, no.~5,
  pp.1007--1065, May 2006.
  DOI:\href{https://doi.org/10.1162/neco.2006.18.5.1007}{10.1162/neco.2006.18.5.1007}

\end{thebibliography}
\end{document}